\pdfoutput=1

\documentclass[11pt]{article}

\usepackage{emnlp2021}

\usepackage{times}
\usepackage{latexsym}
\usepackage{soul}
\usepackage{multirow}
\usepackage{anyfontsize}

\usepackage[T1]{fontenc}

\usepackage[utf8]{inputenc}

\usepackage{microtype}

\usepackage{graphicx}
\usepackage{pgfplots}
\usepackage{comment}

\title{Speechformer: Reducing Information Loss in Direct Speech Translation}

\author{Sara Papi \textsuperscript{1,2 \includegraphics[scale=0.1]{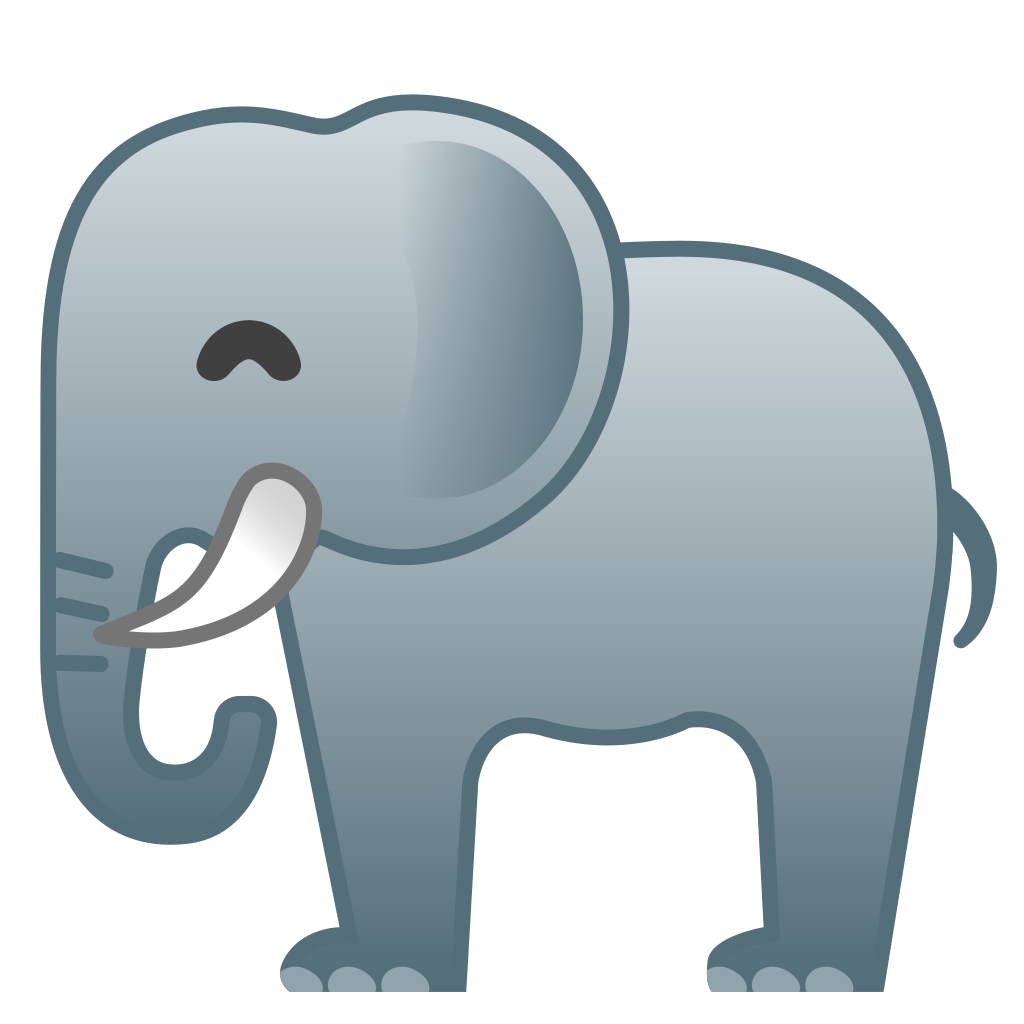}}, Marco Gaido\textsuperscript{1,2 \includegraphics[scale=0.1]{imgs/22246-elephant-icon.png}}, Matteo Negri\textsuperscript{1}, Marco Turchi\textsuperscript{1} \\
\textsuperscript{1}Fondazione Bruno Kessler \\
\textsuperscript{2}University of Trento \\
\texttt{\{spapi,mgaido,negri,turchi\}@fbk.eu}
}

\begin{document}
\maketitle
\begin{abstract}
\newcommand\blfootnote[1]{%
  \begingroup
  \renewcommand\thefootnote{}\footnote{#1}%
  \addtocounter{footnote}{-1}%
  \endgroup
}
\blfootnote{\includegraphics[scale=0.1]{imgs/22246-elephant-icon.png} The authors contributed equally.}
Transformer-based models have gained increasing popularity achieving state-of-the-art performance in many research fields including speech translation.
However, Transformer's quadratic complexity with respect to the input sequence length prevents its adoption \textit{as is} with audio signals, which are typically represented by long sequences.
Current solutions resort to an initial sub-optimal compression based on a fixed sampling of raw audio features.
Therefore, potentially useful linguistic information is not accessible to higher-level layers in the architecture. 
To solve this issue, we propose Speechformer, an architecture that, thanks to a reduced memory usage in the attention layers, avoids the initial lossy compression and aggregates information only at a higher level according to more informed linguistic criteria.
Experiments on three language pairs (en$\rightarrow$de/es/nl) show the efficacy of our solution, with gains of up to 0.8 BLEU on the standard MuST-C corpus and of up to 4.0 BLEU in a low resource scenario.
\end{abstract}

\section{Introduction}
Speech-to-text translation (ST) has been traditionally approached with cascade architectures consisting of a pipeline of two sub-components \cite{StentifordSteer88,Waibel1991b}: an automatic speech recognition (ASR), 
which transforms the audio input into a textual representation, and a machine translation (MT) model, which projects the transcript into the target language.
A more recent approach consists in directly translating speech into target text using a single model \cite{berard_2016,weiss2017sequence}.
This direct solution has interesting advantages 
\cite{sperber-paulik-2020-speech}:
\emph{i)} it can better exploit audio information (e.g. prosody) during the translation phase, \emph{ii)} it has lower latency, and \emph{iii)} it is not affected by error propagation. 
Thanks to these advantages, the initially huge performance gap with cascade systems has gradually closed \cite{ansari-etal-2020-findings}, motivating research towards further improvements.

Direct ST models are fed with features extracted from the audio 
with high frequency (usually every 10ms).
This, on average, makes the resulting input sequence of vectors $\sim$10 times longer than the corresponding text,
leading to an intrinsically redundant (i.e. long and repetitive) representation.
For this reason, it is not possible to process speech data with a vanilla Transformer encoder \cite{transformer}, whose self-attention layers have quadratic memory complexity with respect to the input length.
State-of-the-art architectures tackle the problem by collapsing adjacent vectors in a fixed way, i.e. by mapping a predefined number of vectors (usually 4) into a single one, either using strided convolutional layers \cite{berard_2018,di-gangi-etal-2019-enhancing,wang-etal-2020-fairseq} or by stacking them \cite{sak2015fast}.
As a \textit{positive} side effect, these length reduction solutions
lower input redundancy. 
As a \textit{negative} side effect,
they disregard the variability over time of the amount of linguistic and phonetic information in audio signals (e.g. due to pauses and speaking rate variations) by giving equal weight to all features.
In doing this, relevant features are penalized and considered equally important to the irrelevant ones, resulting in an information loss.

Recently, \citet{salesky-etal-2019-exploring-ph} obtained considerable translation quality gains by collapsing consecutive vectors with the same phonetic content instead of compressing them in a fixed way.
\citet{zhang-etal-2020-adaptive} also showed that selecting a small percentage ($\sim$16\%) of input time steps based on their informativeness improves ST quality. 
On the downside, these approaches respectively require adding a model that performs phoneme classification and a pre-trained adaptive feature selection layer on top of an ASR encoder, losing the compactness of direct solutions at the risk of error propagation.

In direct ST, \citet{liu2020bridging} and \citet{gaido-etal-2021-ctc} addressed the problem with a transcript/phoneme-based compression leveraging Connectionist Temporal Classification (CTC -- \citealt{Graves2006ConnectionistTC}).
However, since these methods are applied to the representation encoded by Transformer layers, the initial content-unaware downsampling of the input is still required for memory reasons, at the risk of losing important information.

To avoid initial fixed compression, we propose Speechformer: the first Transformer-based architecture that processes full audio content maintaining the original dimensions
of the input sequence. Inspired by recent work on reducing the memory complexity of the attention mechanism \cite{wang2020linformer}, 
we introduce a novel attention layer -- the ConvAttention -- whose memory requirements are reduced by means of convolutional layers.
As the benefits of avoiding the initial lossy compression might be outweighed by the increased redundancy of the encoded audio features, we aggregate the high-level representation of the input sequence in a linguistically informed way,
as in \cite{liu2020bridging,gaido-etal-2021-ctc}. In other words, we collapse vectors representing the same linguistic atomic content (words, sub-words, pauses)
into a single element, since they express the same linguistic information.
The usage of the ConvAttention and of the linguistically motivated compression
produces a considerably shorter, yet informative, sequence that fits the memory requirements of vanilla Transformer encoder layers.
Experiments on three language directions (en$\rightarrow$de/es/nl)
show that the proposed architecture outperforms a state-of-the-art ST model by up to 0.8 BLEU points on the standard MuST-C corpus and obtains significantly larger gains (up to 4.0 BLEU) in a low resource setting where the amount of training data is reduced to 100 hours.

\section{Model}
\label{sec:model}

In this section, we first introduce a novel attention layer that enables to process raw audio features without downsampling ($\S\ref{sec:convatt}$).
Then, we present an architecture that leverages this attention mechanism in the first encoder layers and reduces the redundancy of the more informative but longer resulting sequences with CTC compression ($\S\ref{sec:speechformer}$).

\begin{figure}[tbp]
\centering
\includegraphics[width=5cm]{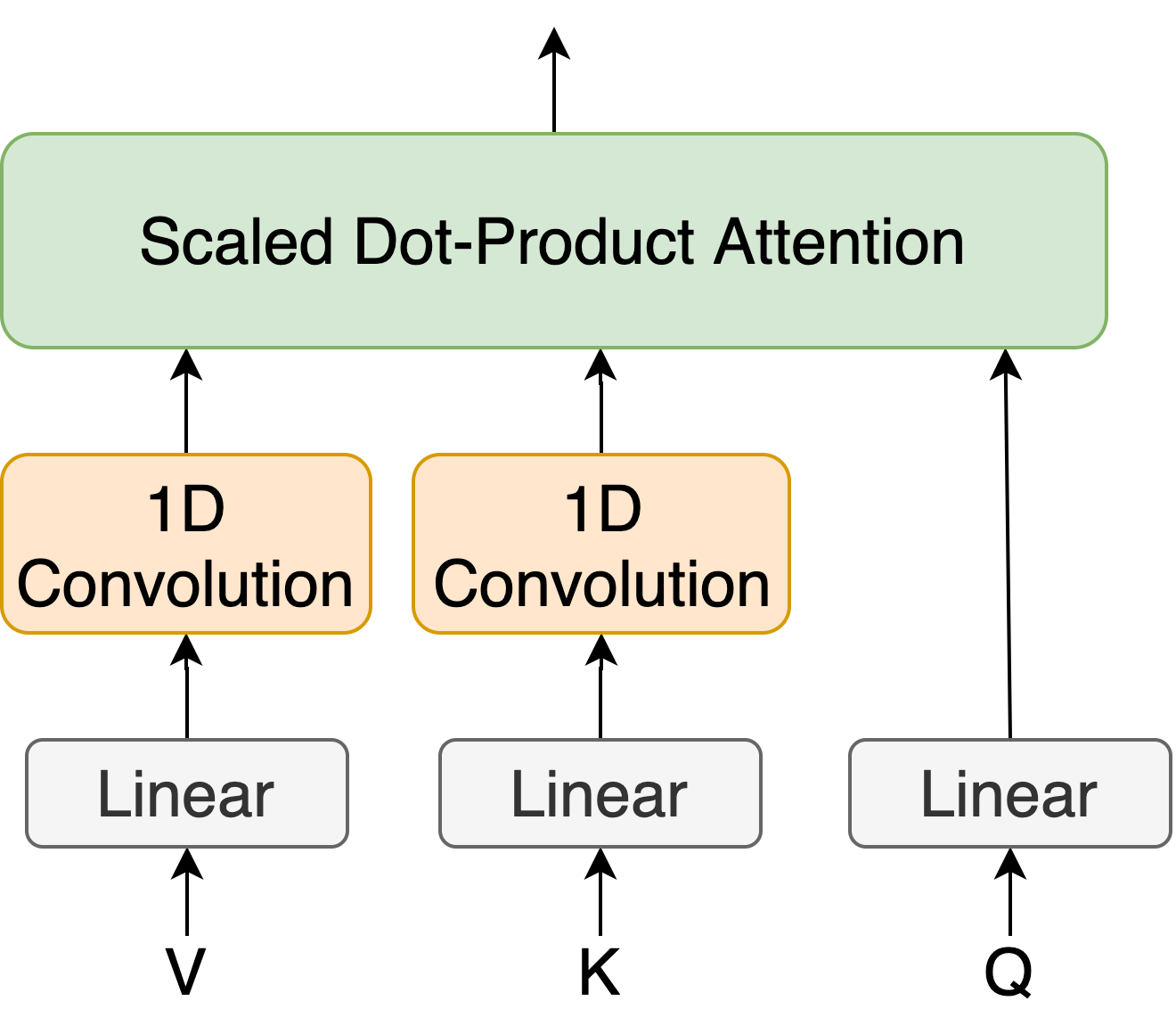}
\caption{Attention mechanism with the proposed convolutional compression of \textit{K} and \textit{V}.
}
\label{fig:attn}
\end{figure}

\subsection{ConvAttention layer}
\label{sec:convatt}

State-of-the-art ST models employ convolutional neural networks to sample the feature sequence to a lower dimension (typically by a factor of 4), enabling the use of Transformer layers otherwise impossible given their memory consumption. 
Outside ST, the Linformer architecture \cite{wang2020linformer} has been recently proposed to reduce
the quadratic complexity of the product between the attention matrix (resulting from the product of the query -- \textit{Q} -- and key -- \textit{K} -- matrices) and the value (\textit{V}) matrix by applying a linear projection to \textit{K} and \textit{V}.
These projections bring the dimension of the sequence length of \textit{K} and \textit{V} to a fixed value, yielding a linear memory complexity.
However, a direct application of this architecture to ST is problematic due to the high variability in audio lengths. 
On one side, mapping those sequences to a fixed dimension can cause an excessive information loss, with a consequent performance drop. 
On the other, it poses technical issues: the linear projection matrix has size $n \times k$, where $n$ is the maximum input length and $k$ is the fixed dimension.
If the input has a length $n'$ shorter than $n$, which is a common case due to the high variability in length of audio sequences, only the first $n'$ weights of the matrix are updated. This results in gradients of different dimensions across GPUs, leading to training failures due to inconsistencies.

To avoid the aforementioned problems, we propose the adoption of ConvAttention (Figure \ref{fig:attn}), in which the linear projections of the Linformer architecture are substituted, both in $K$ and $V$, with a single 1D convolutional layer.
Hence, the length of the sequences used in the scaled dot-product attention depends on the stride of the convolution, a hyper-parameter we named \textit{compression factor} ($\chi$), which controls the memory complexity of the ConvAttention.
Namely, being \textit{n} the temporal dimension of \textit{K} and \textit{V}, the convolution output length is $\frac{n}{\chi}$ and
the complexity of the ConvAttention is $O((\frac{n}{\chi})^2)$, i.e. a $\frac{1}{\chi^2}$ factor lower than a vanilla Transformer self-attention. 
For instance, setting $\chi$ to 4 leads to 
the same memory consumption
as standard ST models with an initial $\times$4 subsampling (i.e. with two initial convolutional layers with stride 2).

Notice that the output sequence length is still equal to the input sequence length as it depends on the length of $Q$ that is not modified.

\subsection{Speechformer}
\label{sec:speechformer}
The introduction of ConvAttention layers allows us to avoid 
sub-optimal fixed compressions that disregard the variability over time in the amount of audio information.
However, since an encoder 
consisting only of
ConvAttention layers does not compress the length of the original input sequence, the decoder will be fed with long and redundant sequences that are difficult to attend, leading to potential performance degradation.

To overcome this problem, as in \citep{liu2020bridging,gaido-etal-2021-ctc}, we apply a content-informed compression to high-level hidden states trained using the CTC loss \cite{Graves2006ConnectionistTC} to represent the linguistic content.
Specifically, the CTC loss produces a prediction for each input time step and then merges equal predictions for consecutive time steps.
The resulting sequence is compared with the reference, which is the sequence of subwords representing the transcript of the input utterance. 
CTC compression, similarly to the loss computation, collapses consecutive features corresponding to the same predictions, averaging them.
After this operation, the sequence is reduced to a representation dimensionally closer to its textual content, which can be processed by the original attention mechanism without the need of approximations.

\begin{figure}[tb]
\hspace*{1.5cm}
\includegraphics[width=3.7cm]{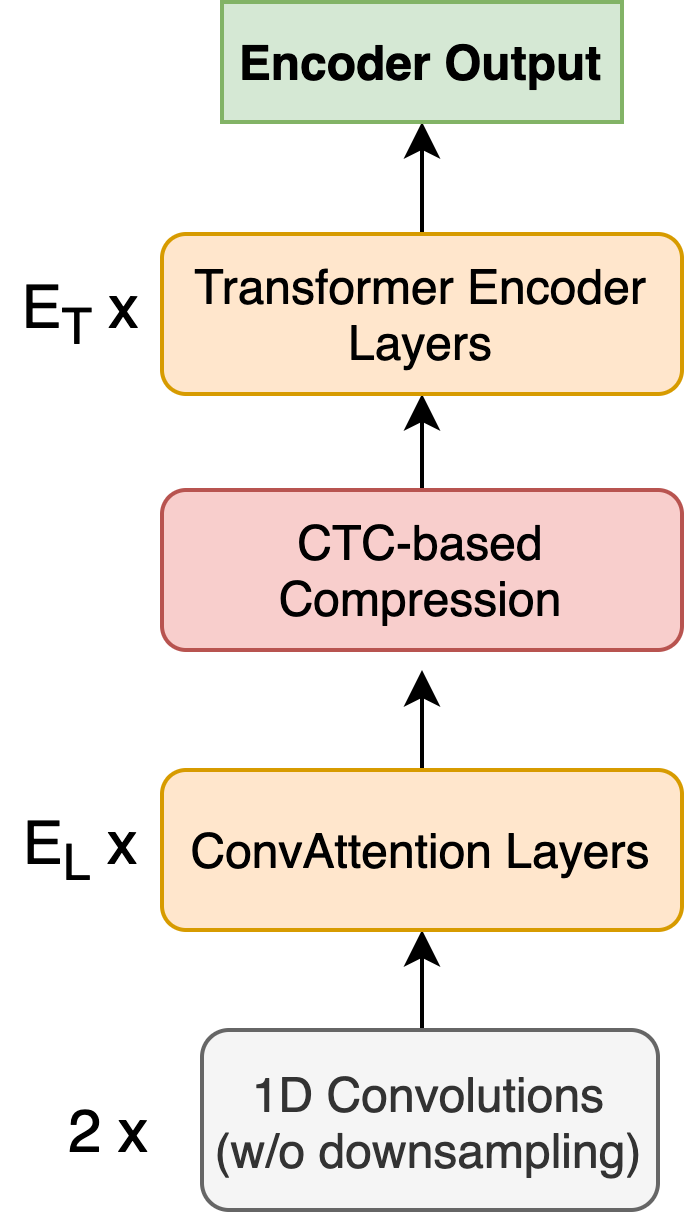}
\caption{Speechformer architecture with \textit{$E_{L}$} ConvAttention Layers and \textit{$E_{T}$} Transformer Encoder Layers.}
\label{fig:hybrid}
\end{figure}

Speechformer (see Figure \ref{fig:hybrid}), is composed of $E_{L}$ ConvAttention layers up to a CTC compression layer, after which there are $E_{T}$ Transformer encoder layers.
The $E_{L}$ ConvAttention layers are meant to learn the linguistic content of the input audio while the $E_{T}$ Transformer encoder layers are in charge of learning higher-level semantic representations, i.e. the encoder outputs, which the decoder has to convert into a text in the target language.
We also maintain the two 1D convolutional layers before the ConvAttention layers but without striding, so that no sub-sampling is applied to the input. We make this choice both to keep the number of parameters comparable to the existing architectures, and to let the model learn a better representation of the input before feeding it to the attention mechanism.

\begin{table}[tb]
\centering
\small
\begin{tabular}{c|ccccc}
 \hline
 kernel & 16 & 16 & 8 & 8 & 4\\
 $\chi$  & 16 & 8 & 8 & 4 & 4 \\
 \hline
 BLEU & 19.7 & 20.6 & 20.5 & \textbf{21.3} & 20.2 \\
\hline
\end{tabular}
\caption{BLEU on MuST-C en-de dev set varying the compression factor $\chi$ and 1D convolutional kernel size. The scores are obtained without label smoothing.}
\label{tab:parameters}
\end{table}

\begin{table*}[t]
\setlength{\tabcolsep}{2.5pt}
\centering
\small
\begin{tabular}{l|cccc|cccc|cccc|c}
 \hline
 \textbf{Model} & 
\multicolumn{4}{c|}{\textbf{en-de}} &  
  \multicolumn{4}{c|}{\textbf{en-es}} & 
  \multicolumn{4}{c|}{\textbf{en-nl}} &
  \multirow{2}{1.4cm}{\textbf{Inference Time}}
 \\ & 
\multicolumn{2}{c}{dev} &  
  \multicolumn{2}{c|}{tst-COMMON} & 
  \multicolumn{2}{c}{dev} &  
  \multicolumn{2}{c|}{tst-COMMON} & 
  \multicolumn{2}{c}{dev} &  
  \multicolumn{2}{c|}{tst-COMMON}
 \\
 \hline
 \cite{inaguma-etal-2020-espnet} & - & & 22.9 & & - & & 28.0 & & - & & 27.4 & & - \\
 \cite{wang-etal-2020-fairseq} & - & & 22.7 & & - & & 27.2 & & - & & 27.3 & & - \\
 \hline
 Our baseline & 22.5 & & 22.8 & & 31.2 & & 27.9 & & 24.2 & & 27.2 &  & 1.0x \\
 \hspace{0.3cm} + compression & 22.3 & -0.2 & 22.8 & +0.0 & 31.1 & -0.1 & 27.9 & +0.0 & 24.2 & +0.0 & 27.0 & -0.2 & 0.9x \\
 \hline
 Plain ConvAttention & 23.1$^{*}$ & +0.6 & 23.2 & +0.4 & 31.5 & +0.3 & 27.7 & -0.2 & 24.8$^{*}$ & +0.6 & 26.9 & -0.3 & 1.8x \\
 Speechformer & \textbf{23.3$^{*}$} & \textbf{+0.8} & \textbf{23.6$^{*}$} & \textbf{+0.8} & \textbf{31.8$^{*}$} & \textbf{+0.6} &\textbf{28.5$^{*}$} & \textbf{+0.6} & \textbf{24.9$^{*}$} & \textbf{+0.7} & \textbf{27.7$^{*}$} & \textbf{+0.5} & 1.3x \\
 \hline
\end{tabular}
\caption{BLEU score (average over 3 runs) on English$\rightarrow$Dutch (en-nl), English$\rightarrow$German (en-de), and English$\rightarrow$Spanish (en-es) of MuST-C tst-COMMON (tst) and the dev (validation) set. The $^{*}$ symbol indicates statistically significant improvements over the baseline. Statistical significance is computed with a t-test \cite{10.2307/2331554}, 
whose null hypothesis is that the mean of the considered experiment is not higher than the mean of the baseline. We consider the result statistically significant if we can reject the null hypothesis with 95\% confidence.}
\label{tab:results}
\end{table*}

\section{Experimental Settings}

We experimented on three languages of MuST-C \cite{MuST-Cjournal}: English-German, English-Spanish, and English-Dutch. To ensure the reproducibility of our work, all training details are provided in the Appendix and the code -- based on fairseq \cite{ott-etal-2019-fairseq} -- is released open source.\footnote{\url{https://github.com/sarapapi/FBK-fairseq/tree/speechformer_emnlp2021}.}

Following \cite{wang2020linformer}, we share the convolution parameters of the ConvAttention layers both among $K$ and $V$ and among the attention heads.
We select the compression factor and the 1D convolution kernel size
with a set of preliminary experiments on the en-de validation set.  
The compression factor ($\chi$) is chosen among 4, 8, and 16, since 4 is the minimum value 
that avoids out-of-memory 
issues. The kernel size is set either equal to or twice as the value of $\chi$.
Table~\ref{tab:parameters} shows that the combination of a compression factor of 4 and a kernel size of 8 leads to better performance compared to the other combinations. Consequently, in all our experiments we use this setting.

We initialize the ConvAttention weights of Speechformer with those of a pre-trained ST model having only ConvAttention layers in the encoder, since, in the initial random state, the CTC-based compression might not properly reduce the input sequence, leading to out-of-memory issues in the following Transformer encoder layers.
Notice that the pre-training does not improve performance. Indeed, \citet{gaido-etal-2021-ctc} already showed that the encoder pre-training improves the baseline performance only without the additional CTC loss and that the results obtained by training without  CTC loss and with encoder pre-training are identical to those achieved with the additional CTC loss.
These findings have been confirmed in our experiments: \textit{i)} initializing the encoder of the baseline with either an ASR or an ST encoder did not bring any improvement, and \textit{ii)} our results are on par with those obtained with encoder pre-training and no additional CTC loss. We do not include the results with encoder pre-training of the baselines, as they do not bring any additional insight.

\section{Results}
We compare our proposed model to a strong \emph{baseline} represented by a Transformer-based model with initial fixed sub-sampling \cite{wang-etal-2020-fairseq} and its \emph{baseline+compression} variant that includes the average CTC compression strategy, as per \cite{gaido-etal-2021-ctc}. We choose to also develop the second baseline to make the comparison with Speechformer fair since they both use the CTC compression strategy.
Table \ref{tab:results} reports the results computed with SacreBLEU\footnote{\fontsize{8.75}{9}{\texttt{BLEU+c.mixed+\#.1+s.exp+tok.13a+v.1.5.0}}} \cite{post-2018-call}. For each experiment, we report the average over 3 runs to ensure that performance differences do not depend on the fluctuations of particularly good
or bad runs.

First, it can be noticed that our baseline is in line with state-of-the-art architectures trained only on MuST-C \cite{wang-etal-2020-fairseq, inaguma-etal-2020-espnet}. Second, the addition of CTC compression to the baseline model does not bring benefits. This confirms the findings of \newcite{gaido-etal-2021-ctc}, who showed that applying CTC compression using transcripts produces differences in score that are not statistically significant.
Speechformer, instead, results in statistically significant improvements over the baseline in all language directions, with BLEU gains ranging from 0.5 (for en-nl) to 0.8 (for en-de).
As the CTC compression is not helpful for the baseline, we also evaluate a model (\emph{Plain ConvAttention}) whose encoder is a stack of ConvAttention layers, i.e. without vanilla Transformer-encoder layers and any form of compression. The drop in performance with respect to Speechformer varies between 0.4 and 0.8 BLEU on all language pairs, supporting our hypothesis that a non-compressed encoder output is too redundant to be effectively attended by the decoder.

\begin{figure}[th]
\centering
\begin{tikzpicture}[scale=1]
    \footnotesize
    \begin{axis}[
            ybar,
            bar width=.4cm,
            legend style={
                at={(0.19,1)}, 
                anchor=north, font=\footnotesize},
            legend cell align={left},
            tickwidth         = 0pt,
            enlarge x limits  = 0.25,
            symbolic x coords={100h,200h,385h,478h},
            xtick=data,
            ylabel={BLEU},
            ylabel near ticks,
            height=6.5cm,
            width=8.25cm,
        ]
        \addplot table[x=Hours,y=baseline]{\eshours};
        \addplot table[x=Hours,y=Speechformer]{\eshours};
        \addplot table[x=Hours,y=plain]{\eshours};
        \legend{baseline, 
        Speechformer, ConvAttention}
    \end{axis}
\end{tikzpicture}
\caption{\label{fig:es_hours} Architecture comparison varying the amount of en-es training data (478h, 385h, 200h, and 100h).}
\end{figure}
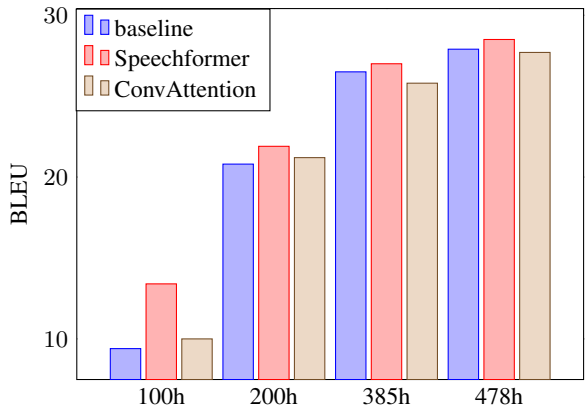

\noindent \textbf{Low-Resource Settings.}
We suppose that the higher gains on en-de may be related to the size of the training data. Indeed, the en-de section of MuST-C used for training is the smallest one,
containing 20\% fewer data than the en-es section and 10\% less than the en-nl one.
Thus, we study Speechformer's performance in different data conditions by progressively reducing the amount of training data.
For this analysis, we select the en-es section of MuST-C as it contains the highest number of hours (478h) among the three languages, and we experiment with three subsets, respectively containing 385h (corresponding to the amount of training data for en-de), 200h, and 100h (which can be considered a limited quantity given that the number of hours is respectively less than half and one fourth of the available data).
Figure \ref{fig:es_hours} shows that the gains obtained by Speechformer over the baseline do not vary significantly between 385h and 478h (0.5 vs 0.6 BLEU).
We can then conclude that the gain variation between en-de and en-es does not depend on the smaller size of the en-de training set. However, in the low resource settings (200h and 100h), the gains obtained by the Speechformer are much larger, amounting to 1.1 BLEU with 200h and 4.0 BLEU with 100h. To validate the robustness of these results, we also experimented on the en-de language pair and obtained consistent results: 
Speechformer outperforms the baseline by 1.5 BLEU (19.6 vs 18.1 BLEU) with 200h of training data and by 1.9 BLEU (9.7 vs 7.8 BLEU) with 100h of training data, achieving a considerable relative improvement of more than 24\%.
Although it brings consistent and significant gains in higher resource scenarios, these experiments show that Speechformer is particularly fruitful in low-resource settings.
We leave to future work the assessment of the behavior of Speechformer in unrestricted data conditions (e.g. when using large ASR corpora to generate pseudo-labelled ST training data).

\pgfplotstableread[row sep=\\,col sep=&]{
    Hours & baseline & baselinecompr & plain & Speechformer \\
    478h      & 27.9 & 27.9 & 27.7 & 28.5 \\
    385h      & 26.5 & 26.6 & 25.8 & 27.0 \\
    200h      & 20.8 & 20.4 & 21.2 & 21.9 \\
    100h      & 9.4 &  & 10.0 & 13.4 \\
}\eshours

\noindent \textbf{Inference Time.} 
The ConvAttention layers process the whole input sequences, which are 4 times larger than those elaborated by the baseline attention mechanism. Thereby, a slow-down at inference time is expected, especially for the \emph{Plain ConvAttention}, whose encoder layers are all ConvAttention layers.
The last column of Table \ref{tab:results} confirms that the \emph{Plain ConvAttention} architecture is 1.8 times slower than the baseline, i.e. the inference time is nearly twice. 
Speechformer is also
slower than the baseline, but the overhead amounts to only 30\% instead of 80\%.
Moreover, it can be noticed that the size of the attention matrix -- and therefore the corresponding computational cost -- can be controlled in the Speechformer with the \textit{compression factor} ($\chi$) hyper-parameter.
We leave to future studies the analysis of the trade-off between overall translation quality and inference time, which is usually irrelevant in offline ST, but becomes critical in simultaneous scenarios.

\noindent \textbf{Manual Analysis.}
Lastly, we inspected the baseline and Speechformer outputs to better understand the reason behind the improvements brought by our architecture.
This qualitative analysis was conducted on a sample of 200 sentences of the en-de test set -- the language direction showing the largest gap between the systems (+0.8, see Table \ref{tab:results}) -- by a professional linguist with C2 German level.
It emerged (see the Appendix for examples) that Speechformer tends to have better word-ordering, a typical problem arising when translating from an SVO language like English to an SOV language like German.
Furthermore, Speechformer outputs display a better punctuation positioning -- attributable to an improved
handling of pauses and prosody -- and a reduction of the number of audio misunderstandings and omissions.
Together with the overall BLEU gains, these findings provide us with interesting hints about the potential of Speechformer.

\section{Conclusion}
In the wake of previous works showing the benefits of a content-informed compression over fixed downsampling of the audio features, we proposed Speechformer: the first ST Transformer-based model able to encode the whole raw audio features without any sub-optimal initial subsampling typical of current state-of-the-art models.
Our solution is made possible by the introduction of a modified attention mechanism -- the ConvAttention -- that reduces the memory complexity to $O((\frac{n}{\chi})^2)$.
As the plain application of ConvAttention layers leads to redundant sequences, high-level hidden states are compressed with a CTC-based strategy to obtain a compact, yet informative representation that can be processed by vanilla Transformer encoder layers.
Experiments on three language pairs show that Speechformer significantly outperforms a state-of-the-art ST model by 0.5-0.8 BLEU, reaching a peak of 4 BLEU points in a low resource scenario.

\section*{Acknowledgements}

The authors would like to thank Alina Karakanta for the support and help with the manual analysis.

\bibliography{anthology,custom}
\bibliographystyle{acl_natbib}

\appendix

\section{Training Details}
\label{sec:training}

All our models are composed by 12 encoder layers and 6 decoder layers with 8 attention heads and are trained using label smoothed cross entropy \cite{szegedy2016rethinking} with the auxiliary CTC loss \cite{kim-et-al-2017-joint,bahar2019comparative} and Adam optimizer \cite{DBLP:journals/corr/KingmaB14}.
The number of parameters is $\sim$77M for the baseline and $\sim$79M for the Speechformer.
The CTC is computed at the 8th encoder layer and its role is to predict the source transcription (lowercased and without punctuation), as in \cite{liu2020bridging}. 
The learning rate is set to 1e-3 with an inverse square-root scheduler and 10,000 warm-up updates.
Mini-batches contain up to 5,000 tokens and we update gradients every 16 mini-batches.
We apply SpecAugment \cite{Park_2019} and utterance-level cepstral mean and variance normalization. We filter out samples with duration exceeding 30s. The text is segmented in sub-word units with transcript and target Sentencepiece \cite{kudo-richardson-2018-sentencepiece} unigram language models \cite{kudo-2018-subword} with size 5,000 and 8,000 respectively.
We average 7 checkpoints around the best on the validation loss. Trainings were performed with 4 GPUs NVIDIA Tesla K-80 with 12GB of RAM and lasted about 3 days.

{
\begin{table*}[h!]
  \centering
  \setlength{\tabcolsep}{1.5pt}
  \small
  \begin{tabular}{p{2cm}|p{13cm}}
      \hline
      \multicolumn{2}{c}{(a) Word ordering} \\
      \hline
      \textbf{Audio} & It was a way that parents could figure out which were the right public \textbf{schools} for their \textbf{kids}. \\
      \textbf{Reference} & Es ging um eine Methode, mit der Eltern herausfinden können, welche die richtigen öffentlichen \textbf{Schulen} für ihre \textbf{Kinder} sind. \\
      \textbf{Baseline} & Es war eine Möglichkeit, dass Eltern herausfinden konnten, welche für ihre \textbf{Kinder} die richtige öffentliche \textbf{Schule} war.
      
      \textit{It was an opportunity for the parents to find out which were for their \textbf{children} the right public \textbf{schools}.}
      \\
      \textbf{Speechformer} & Es war eine Methode, mit der Eltern herausfinden konnten, welche die richtigen öffentlichen \textbf{Schulen} für ihre \textbf{Kinder} waren.
      
      \textit{It was a method with which the parents could find out which were the right public \textbf{schools} for their \textbf{children}.}
      \\
      \hline
  \multicolumn{2}{c}{(b) Punctuation handling} \\
      \hline
      \textbf{Audio} & \textbf{So, sir, can you help me?} I need help. \\
      \textbf{Reference} & \textbf{Also, mein Herr, können Sie mir helfen?} Ich brauche Hilfe. \\
      \textbf{Baseline} & Es ist also möglich, mir zu helfen.
      
      \textit{So it is possible to help me.}
      \\
      \textbf{Speechformer} & \textbf{Also, können Sie mir helfen?} Ich habe keine Hilfe.
      
      \textit{\textbf{So can you help me?} I have no help.}
      \\
      \hline
      \multicolumn{2}{c}{(c) Audio misunderstanding} \\
      \hline
      \textbf{Audio} & You see Aluminum was the most valuable metal on the Planet, worth more than Gold and \textbf{Platinum}. \\
      \textbf{Reference} & Aluminium war zu dieser Zeit das wertvollste Metall auf dem Planeten, wertvoller als Gold und \textbf{Platin}. \\
      \textbf{Baseline} & Aluminium war die wertvollste Metallart auf dem Planeten, mehr als Gold und \textbf{Pflanzen}.
      
      \textit{Aluminum was the most valuable type of metal on the planet, more than gold and \textbf{plants}.}
      \\
      \textbf{Speechformer} & Aluminium war das wertvollste Metall auf dem Planeten, mehr als Gold und \textbf{Platin}.
      
      \textit{
Aluminum was the most valuable metal on the planet, more than gold and \textbf{platinum}.}
      \\
      \hline
      \multicolumn{2}{c}{(d) Omission} \\
      \hline
      \textbf{Audio} & But the amazing thing about cities is they're worth so much more than \textbf{it costs} to build them. \\
      \textbf{Reference} & Aber das Erstaunliche an Städten ist, dass sie so viel mehr wert sind, als \textbf{es kostet} sie zu bauen. \\
      \textbf{Baseline} & Aber das Faszinierende an Städten ist, dass es viel mehr wert ist, als es zu bauen. 
      
      \textit{But the fascinating thing about cities is that it's worth a lot more than building it.}
      \\
      \textbf{Speechformer} & Aber das Erstaunliche an Städten ist, dass sie viel mehr wert sind als sie \textbf{es kostet}, sie zu bauen.
      
      \textit{
But the amazing thing about cities is that they are worth a lot more than \textbf{it costs} to build them.}
      \\
      \hline
  \end{tabular} 
  \caption{\label{tab:examples} Examples of translation problems -- \textit{(a)}, \textit{(b)}, \textit{(c)} -- and omissions -- \textit{(d)} -- that Speechformer does not suffer from while baseline does.}
\end{table*}
}

\section{Output examples}

Table~\ref{tab:examples} provides examples of the German translations generated by the baseline and by Speechformer for four utterances of the MuST-C test set, chosen among 200 sentences manually inspected by a C2 German speaker. These sentences have been selected to highlight the specific aspects that are better handled by Speechformer.

Example \emph{(a)} exhibits a wrong word ordering present in the baseline output, i.e. it anticipates ``für ihre Kinder'' (\textit{for their kids}) with respect to ``die richtigen öffentlichen Schulen'' (\textit{the right public schools}). 
Our proposed architecture, instead, translates the sentence in the correct order, making the translation easier to be read and understood.

Example \emph{(b)} displays that Speechformer shows better punctuation handling, which -- we hypothesize -- is the result of an improved representation of prosody and pauses. In this example, for instance, our architecture is capable of detecting a question (i.e. \textit{So can you help me?}) and translating it, while the baseline does not translate the input in question form and omits the last part of the audio content. Listening to the audio, we noticed a long pause after the question. We suppose that this
pause led the baseline to conclude the sentence, while Speechformer managed to translate the remaining part of the utterance by going beyond that pause.

Our architecture
shows an improved encoding of audio features that is reflected in its superior understanding of audio content. This emerges, indeed, from example \emph{(c)}, where the word \textit{Platinum} is correctly recognized and translated by our system, while the baseline misunderstands and translates it in another word, ``Pflanzen'' (\textit{plants}), with a completely different meaning.
The better audio understanding of Speechformer is present in example \emph{(d)} as well.
On the contrary, the baseline omits part of the original sentence (i.e. \textit{it costs}), with a huge impact on the meaning of the resulting sentence, while Speechformer does not lose audio details and produces a complete translation.
In this example, we can also notice that our system better solves pronominal resolution as it chooses \emph{sie}, which follows the grammatical gender and number of \emph{Staedten} (i.e. plural feminine), while the baseline uses \emph{es}, which wrongly agrees with \emph{das Faszinierende} (i.e. singular neuter).


\section{Effect of Label Smoothing}
Label smoothing \cite{szegedy2016rethinking} is a widely adopted regularization factor \cite{zhang-2021-delving-deep}. As such, a more complex architecture that processes longer and potentially more redundant inputs -- like our proposed Speechformer -- can benefit more from its adoption. Hence, to validate that our gains are not due to a better regularization of the models and to assess the effect of label smoothing, we run experiments using the cross entropy loss without smoothing factor. The results are reported in Table \ref{tab:results_ls}. Compared with the scores reported in Section 4 of the paper, we can see that label smoothing brings significant gains for all the systems (ranging from 1.5 to 2.0 BLEU points). 
Most importantly, the improvements of the Speechformer architecture (0.5-1.1 BLEU) are similar to those achieved with label smoothing (0.5-0.8 BLEU). 
The minimal difference can be explained by statistical variations of the results, considering that those obtained without label smoothing are computed on a single run. We can conclude that these results confirm the efficacy of our architecture and the validity of our experiments, showing that they are not biased by a higher regularization that might favor our solution over the baseline.

\begin{table}[t]
\setlength{\tabcolsep}{2.5pt}
\centering
\small
\begin{tabular}{l|cc|cc|cc}
 \hline
 \textbf{Model} & 
\multicolumn{2}{c|}{\textbf{en-de}} &  
  \multicolumn{2}{c|}{\textbf{en-es}} & 
  \multicolumn{2}{c}{\textbf{en-nl}} 
 \\
 \hline
 baseline & 21.2 & & 26.2 & & 25.5 & \\
 \hspace{0.3cm} + compression & 21.2 & +0.0 & 26.0 & -0.2 & 25.1 & -0.4 \\
 \hline
 Plain ConvAttention & 21.6 & +0.4 & 25.6$^{*}$ & -0.6 & 25.6 & +0.1 \\
 Speechformer & \textbf{22.3$^{*}$} & \textbf{+1.1} & \textbf{26.7$^{*}$} & \textbf{+0.5} & \textbf{26.2$^{*}$} & \textbf{+0.7} \\
 \hline
\end{tabular}
\caption{BLEU score on three language pairs of MuST-C tst-COMMON. The $^{*}$ symbol indicates statistically significant improvements over the baseline computed with bootstrap resampling \cite{koehn-2004-statistical} with 10,000 samples, 1,000 sample size and 95\% significance level.
}
\label{tab:results_ls}
\end{table}

\end{document}


\appendix

\section{Training Details}
\label{sec:training}

All our models are composed by 12 encoder layers and 6 decoder layers with 8 attention heads and are trained using label smoothed cross entropy \cite{szegedy2016rethinking} with the auxiliary CTC loss \cite{kim-et-al-2017-joint,bahar2019comparative} and Adam optimizer \cite{DBLP:journals/corr/KingmaB14}.
The number of parameters is $\sim$77M for the baseline and $\sim$79M for the Speechformer.
The CTC is computed at the 8th encoder layer and its role is to predict the source transcription (lowercased and without punctuation), as in \cite{liu2020bridging}. 
The learning rate was set to 1e-3 with an inverse square-root scheduler and 10,000 warm-up updates.
Mini-batches contain up to 5,000 tokens and we update gradients every 16 mini-batches.
We apply SpecAugment \cite{Park_2019} and utterance-level cepstral mean and variance normalization. We filtered out samples with duration exceeding 30s. The text is segmented in sub-word units with transcript and target Sentencepiece \cite{kudo-richardson-2018-sentencepiece} unigram language models \cite{kudo-2018-subword} with size
5,000 and 8,000 respectively.
We average 7 checkpoints around the best on the validation loss.
Trainings, lasted about 3 days, were performed with 4 GPUs NVIDIA Tesla K-80 with 12GB of RAM.

\section{Effect of Label Smoothing}
\begin{table}[ht]
\setlength{\tabcolsep}{2.5pt}
\centering
\small
\begin{tabular}{l|cc|cc|cc}
 \hline
 \textbf{Model} & 
\multicolumn{2}{c|}{\textbf{en-de}} &  
  \multicolumn{2}{c|}{\textbf{en-es}} & 
  \multicolumn{2}{c}{\textbf{en-nl}} 
 \\
 \hline
 baseline & 21.2 & & 26.2 & & 25.5 & \\
 \hspace{0.3cm} + compression & 21.2 & +0.0 & 26.0 & -0.2 & 25.1 & -0.4 \\
 \hline
 Plain ConvAttention & 21.6 & +0.4 & 25.6$^{*}$ & -0.6 & 25.6 & +0.1 \\
 Speechformer & \textbf{22.3$^{*}$} & \textbf{+1.1} & \textbf{26.7$^{*}$} & \textbf{+0.5} & \textbf{26.2$^{*}$} & \textbf{+0.7} \\
 \hline
\end{tabular}
\caption{BLEU score on three language pairs of MuST-C tst-COMMON. $^{*}$ indicates statistically significant improvements over the baseline computed with bootstrap resampling \cite{koehn-2004-statistical} with 10,000 samples, 1,000 sample size and 95\% significance level.
}
\label{tab:results_ls}
\end{table}

{
\begin{table*}[h!]
  \centering
  \setlength{\tabcolsep}{1.5pt}
  \small
  \begin{tabular}{p{2cm}|p{13cm}}
      \hline
      \multicolumn{2}{c}{(a) Word ordering} \\
      \hline
      \textbf{Audio} & It was a way that parents could figure out which were the right public \textbf{schools} for their \textbf{kids}. \\
      \textbf{Reference} & Es ging um eine Methode, mit der Eltern herausfinden können, welche die richtigen öffentlichen \textbf{Schulen} für ihre \textbf{Kinder} sind. \\
      \textbf{Baseline} & Es war eine Möglichkeit, dass Eltern herausfinden konnten, welche für ihre \textbf{Kinder} die richtige öffentliche \textbf{Schule} war.
      
      \textit{It was an opportunity for the parents to find out which were for their \textbf{children} the right public \textbf{schools}.}
      \\
      \textbf{Speechformer} & Es war eine Methode, mit der Eltern herausfinden konnten, welche die richtigen öffentlichen \textbf{Schulen} für ihre \textbf{Kinder} waren.
      
      \textit{It was a method with which the parents could find out which were the right public \textbf{schools} for their \textbf{children}.}
      \\
      \hline
      
      
      \multicolumn{2}{c}{(b) Punctuation handling} \\
      \hline
      \textbf{Audio} & \textbf{So, sir, can you help me?} I need help. \\
      \textbf{Reference} & \textbf{Also, mein Herr, können Sie mir helfen?} Ich brauche Hilfe. \\
      \textbf{Baseline} & Es ist also möglich, mir zu helfen.
      
      \textit{So it is possible to help me.}
      \\
      \textbf{Speechformer} & \textbf{Also, können Sie mir helfen?} Ich habe keine Hilfe.
      
      \textit{\textbf{So can you help me?} I have no help.}
      \\
      \hline
      \multicolumn{2}{c}{(c) Audio misunderstanding} \\
      \hline
      \textbf{Audio} & You see Aluminum was the most valuable metal on the Planet, worth more than Gold and \textbf{Platinum}. \\
      \textbf{Reference} & Aluminium war zu dieser Zeit das wertvollste Metall auf dem Planeten, wertvoller als Gold und \textbf{Platin}. \\
      \textbf{Baseline} & Aluminium war die wertvollste Metallart auf dem Planeten, mehr als Gold und \textbf{Pflanzen}.
      
      \textit{Aluminum was the most valuable type of metal on the planet, more than gold and \textbf{plants}.}
      \\
      \textbf{Speechformer} & Aluminium war das wertvollste Metall auf dem Planeten, mehr als Gold und \textbf{Platin}.
      
      \textit{
Aluminum was the most valuable metal on the planet, more than gold and \textbf{platinum}.}
      \\
      \hline
      \multicolumn{2}{c}{(d) Omission} \\
      \hline
      \textbf{Audio} & But the amazing thing about cities is they're worth so much more than \textbf{it costs} to build them. \\
      \textbf{Reference} & Aber das Erstaunliche an Städten ist, dass sie so viel mehr wert sind, als \textbf{es kostet} sie zu bauen. \\
      \textbf{Baseline} & Aber das Faszinierende an Städten ist, dass es viel mehr wert ist, als es zu bauen. 
      
      \textit{But the fascinating thing about cities is that it's worth a lot more than building it.}
      \\
      \textbf{Speechformer} & Aber das Erstaunliche an Städten ist, dass sie viel mehr wert sind als sie \textbf{es kostet}, sie zu bauen.
      
      \textit{
But the amazing thing about cities is that they are worth a lot more than \textbf{it costs} to build them.}
      \\
      \hline
  \end{tabular} 
  \caption{\label{tab:examples} Examples of translation problems -- \textit{(a)}, \textit{(b)}, \textit{(c)} -- and omissions -- \textit{(d)} -- that Speechformer does not suffer from while baseline does.}
\end{table*}
}

\mg{Label smoothing \cite{szegedy2016rethinking} is a widely adopted regularization factor \cite{zhang-2021-delving-deep}. As such, a more complex architecture that processes longer and potentially more redundant inputs -- like our proposed Speechformer -- can gain more from its adoption. Hence, to validate that our gains are not due to a better regularization of the models and to assess the effect of label smoothing, we executed experiments using the cross entropy loss without smoothing factor. The results are reported in Table \ref{tab:results_ls}. Compared with the scores reported in Section 4 of the paper, we can see that: \textit{i)} label smoothing brings significant improvements for all the systems (with gains ranging from 1.5 to 2 BLEU points); \textit{ii)} most importantly, the improvements of the Speechformer architecture (0.5-1.1 BLEU) are similar to those achieved with label smoothing (0.5-0.8 BLEU) and the minimal difference can be explained by statistical variation of the results, considering that the results without label smoothing are reported on a single run. We can conclude that these results confirm the efficacy of our architecture and the validity of our experiments, showing that they are not biased by a higher regularization that might favor our solution over the baseline.}

\section{Output examples}


Table~\ref{tab:examples} provides examples of the German translations generated by the baseline and by Speechformer for four utterances of the MuST-C test set, chosen among 200 sentences manually inspected by a C2 German speaker. These sentences have been selected to highlight the specific aspects that are better handled by our proposed solution.


Example \emph{(a)} exhibits a wrong word ordering 
present in the baseline output, i.e. it anticipates ``für ihre Kinder'' (\textit{for their kids}) with respect to ``die richtigen öffentlichen Schulen'' (\textit{the right public schools}). 
Speechformer, instead, translates the sentence in the correct order, making the translation easier to be read and understood.

Also punctuation handling is improved by our model that, by leveraging the prosody present in the audio, is capable of detecting a question (i.e. \textit{So can you help me?}) and translating it, as shown in example \emph{(b)}. On the contrary, the baseline does not capture these audio characteristics and does not translate the input in question form, besides omitting the last part of the reference sentence.

Speechformer's improved encoding of audio features is also reflected in its superior understanding of audio content. This emerges from example \emph{(c)}, where the word \textit{Platinum} is correctly recognized and translated by our system, while the baseline misunderstands and translates it in another word, ``Pflanzen'' (\textit{plants}),
with a completely different meaning.
The better audio understanding of the Speechformer is
present in example \emph{(d)} as well.
Indeed, the baseline omits part of the original sentence (i.e. \textit{it costs}), with a huge impact on the meaning of the resulting sentence, while Speechformer does not lose audio details and produces a complete translation.
In this example, we can also notice that our system better solves pronominal resolution as it chooses \emph{sie}, which follows the grammatical gender and number of \emph{Staedten} (i.e. plural feminine), while the baseline uses \emph{es}, which wrongly agrees with \emph{das Faszinierende} (i.e. singular neuter).

\bibliography{anthology,custom}
\bibliographystyle{acl_natbib}